\documentclass[sigconf,screen]{acmart}
\AtBeginDocument{%
  \providecommand\BibTeX{{%
    \normalfont B\kern-0.5em{\scshape i\kern-0.25em b}\kern-0.8em\TeX}}}

\setcopyright{acmlicensed}
\acmPrice{15.00}
\acmDOI{10.1145/3445814.3446753}
\acmYear{2021}
\copyrightyear{2021}
\acmSubmissionID{asplos21main-p969-p}
\acmISBN{978-1-4503-8317-2/21/04}
\acmConference[ASPLOS '21]{Proceedings of the 26th ACM International Conference on Architectural Support for Programming Languages and Operating Systems}{April 19--23, 2021}{Virtual, USA}
\acmBooktitle{Proceedings of the 26th ACM International Conference on Architectural Support for Programming Languages and Operating Systems (ASPLOS '21), April 19--23, 2021, Virtual, USA}

\newcommand{\Mod}[1]{\ (\mathrm{mod}\ #1)}
\usepackage{algorithmic}
\usepackage{algorithm}
\usepackage{graphicx}
\usepackage{textcomp}

\usepackage{verbatim}
\usepackage[frozencache=true]{minted}
\usepackage{shortbold}
\usepackage{caption}

\begin{document}

\title{Neural Architecture Search as Program Transformation Exploration}


\author{Jack Turner}
\email{jack.turner@ed.ac.uk}
\affiliation{%
    \institution{University of Edinburgh}
    \country{United Kingdom}
}

\author{Elliot J. Crowley}
\email{elliot.j.crowley@ed.ac.uk}
\affiliation{%
    \institution{University of Edinburgh}
    \country{United Kingdom}
}

\author{Michael  F.P. O'Boyle}
\email{mob@inf.ed.ac.uk}
\affiliation{%
    \institution{University of Edinburgh}
    \country{United Kingdom}
}


\begin{abstract}
    Improving the performance of deep neural networks (DNNs) is important to both the compiler  and neural architecture search (NAS) communities. Compilers apply program transformations in order to  exploit hardware parallelism and memory hierarchy. However, legality concerns mean they fail to exploit the natural robustness of neural networks. In contrast, NAS techniques mutate networks by operations such as the grouping or bottlenecking of convolutions, exploiting the resilience of DNNs.
    In this work, we express  such neural architecture operations as program transformations whose legality depends on a notion of representational capacity. This allows them to be combined with existing  transformations into a unified optimization framework. This unification allows us to express existing NAS operations
    as combinations of simpler transformations. Crucially, it allows us to generate and explore new tensor convolutions. We prototyped the combined framework in TVM and  were able to find optimizations across different DNNs,  that  significantly reduce  inference time - over 3$\times$ in the majority of cases.
    Furthermore, our scheme  dramatically reduces NAS search time. Code is available at~\href{https://github.com/jack-willturner/nas-as-program-transformation-exploration}{this https url}.
\end{abstract}




\maketitle

\section{Introduction}
\label{sec:introduction}

Deep neural networks (DNNs)~\cite{schmidhuber2015deep,lecun2015deep} are everywhere,
and there is  a growing need to implement them efficiently
~\cite{huang2019gpipe, chen2012benchnn, chowdhery2019visual}.
This has led to an explosion in research from application
~\cite{hinton2015distilling} to hardware~\cite{Chen:2014:DSH:2541940.2541967}.

Currently, there are two distinct communities optimizing  DNNs  for commodity devices.
In one, neural architecture  search (NAS) researchers explore different network models
trading off size and accuracy. In the other, compiler developers take the resulting networks
and explore optimzations to deliver hardware performance.  We argue that this  two-stage
deployment process is artificial and can be unified.  This paper shows that program
and neural architecture transformations can  be {\em interleaved}  delivering
significant performance improvement and greater expressivity. Before we describe our approach,
let us briefly look at the two existing communities.

\subsection{Program Transformations}
Within compiler research,
there has been much focus on restructuring
underlying tensor computations %to best utilize hardware
~\cite{Pu:2017:PHS:3132652.3107953, yang2016systematic, cyphers2018intel}.
This involves significant loop nest restructuring~\cite{abadi2016tensorflow,chen2018tvm}
exploiting characteristics of the  target device
e.g.\ vectorization
in SIMD units or memory coalescing in GPUs \cite{Elango:2018:DDL:3211346.3211354}.

There are currently many methods~\cite{ragan2013halide,steuwer2017lift,chen2018tvm,tillet2019triton}
in use.
The polyhedral model
~\cite{vasilache2018tensor,zerrell2019stripe}, in particular, is a natural fit for many standard neural network operations.
A key aspect of a good program  transformation
representation is that sequences of transformations can be easily composed and checked for legality. This allows automatic exploration of a large space of options which can then be evaluated, and the best selected~\cite{steuwer2017lift}.

\subsection{Neural Architecture Search}
The neural architecture search (NAS) community is also concerned with
accuracy, space, and
performance~\cite{elsken2019neural,wistuba2019survey}.
While there have been advances in NAS for language models~\cite{so2019evolved}, large scale studies of NAS tend to focus primarily on convolutional architectures~\cite{Dong2020NAS-Bench-201:,zela2020bench,ying2019bench}.
At the neural architecture level, many techniques have been proposed
for the automatic generation of networks under strict
budgets \cite{he2018amc,yang2018netadapt,wu2019fbnet,tan2019mnasnet}.
These methods focus on overall network
structure, selectively replacing components such as convolutional layers with
computationally cheaper methods to balance the trade-off between accuracy and inference time.

One key strength of NAS is that it leverages the
robustness of neural networks to deformation, reshaping and
transforming them while incurring minimal damage to their
ability to learn.
This relies on networks maintaining their ability to extract feature representations for a given type of input, which we refer to as~\textit{representational capacity}.
If an approximation or compression to a network does not inhibit its ability to learn from data, then it has not damaged the representational capacity of the network.

The ability of neural networks to weather compression without losing representational capacity is well-documented~\cite{denil2013predicting,lecun1989optimal,han2016deep}.
The legality of such neural architecture transformations, however, is not guaranteed and must be evaluated separately through either a training process or a small proxy task.
Furthermore, while powerful,
few NAS techniques explicitly take into account actual hardware behavior, and those
that do face one of three problems.
The first is that they only offer a
black box solution with highly complex methods for predicating the
search process on specific budgets (usually involving reinforcement learning~\cite{he2018amc,tan2019mnasnet}).
The second is that they do so while leaving the compilation
pipeline fixed, often dismissing powerful candidate architectures
because of an inappropriate, fixed choice of
program transformations. The final problem is that they are limited to selecting from a pre-designed list of convolutional alternatives; they cannot synthesize their own.

\subsection{Our Approach}

We have two distinct communities who have the same goal but
are siloed. NAS researchers assume the compiler is a black box bundled
with the hardware, while compiler writers assume that the network
architecture is set in stone. NAS designers can discover good
networks but are limited to pre-defined options; compiler writers can
efficiently exploit hardware structure but miss larger scale
optimization opportunities.

What we want is the best of both worlds. We wish to combine  neural architecture and compiler optimization in a unified framework. In this paper, we recast neural architecture search as program transformation exploration. By including transformations such as grouping and bottlenecking into the compiler optimization space, we leverage both the extensive history of program transformation research, and also discover new forms of neural architecture reduction that would not have been available to us otherwise.

Program transformations are necessarily restricted
as they must be safe. Our solution is to unlock the space of~\textit{neural  transformations} by introducing a new safety metric based on Fisher Potential
~\cite{Turner2020BlockSwap:}. It is a compile-time, cheap-to-compute metric that can reject damaging network changes, {\em eliminating the need to train while searching}.
We, therefore,  judge a transformation to be legal, not by data dependence preservation, but by the preservation of~\textit{representational capacity} (see section \ref{sec:fish}). This unification allows the exploration of a space that leads to more efficient implementations. It also shows that neural architecture options that previously required the engineering efforts of experts to develop, such as spatial bottlenecking, can be expressed as compositions of more fundamental transformations and discovered automatically (see section \ref{sec:spatial}).

Our contributions are as follows:
\begin{enumerate}
    \item We reformulate popular Neural Architecture Search techniques as~\textit{program transformations}.
    \item We use Fisher Potential to  provide transformation safety guarantees without the need to train.
    \item We unify the transformation and architecture search spaces, discovering new types of convolution.
    \item We evaluate 3 networks using these operations, ResNet, ResNext and DenseNet, on 4 platforms and
     demonstrate, in most cases, more than $3\times$ inference speedup over a TVM baseline.
\end{enumerate}

\begin{figure*}[t]
    \centering
    \includegraphics[width=\linewidth]{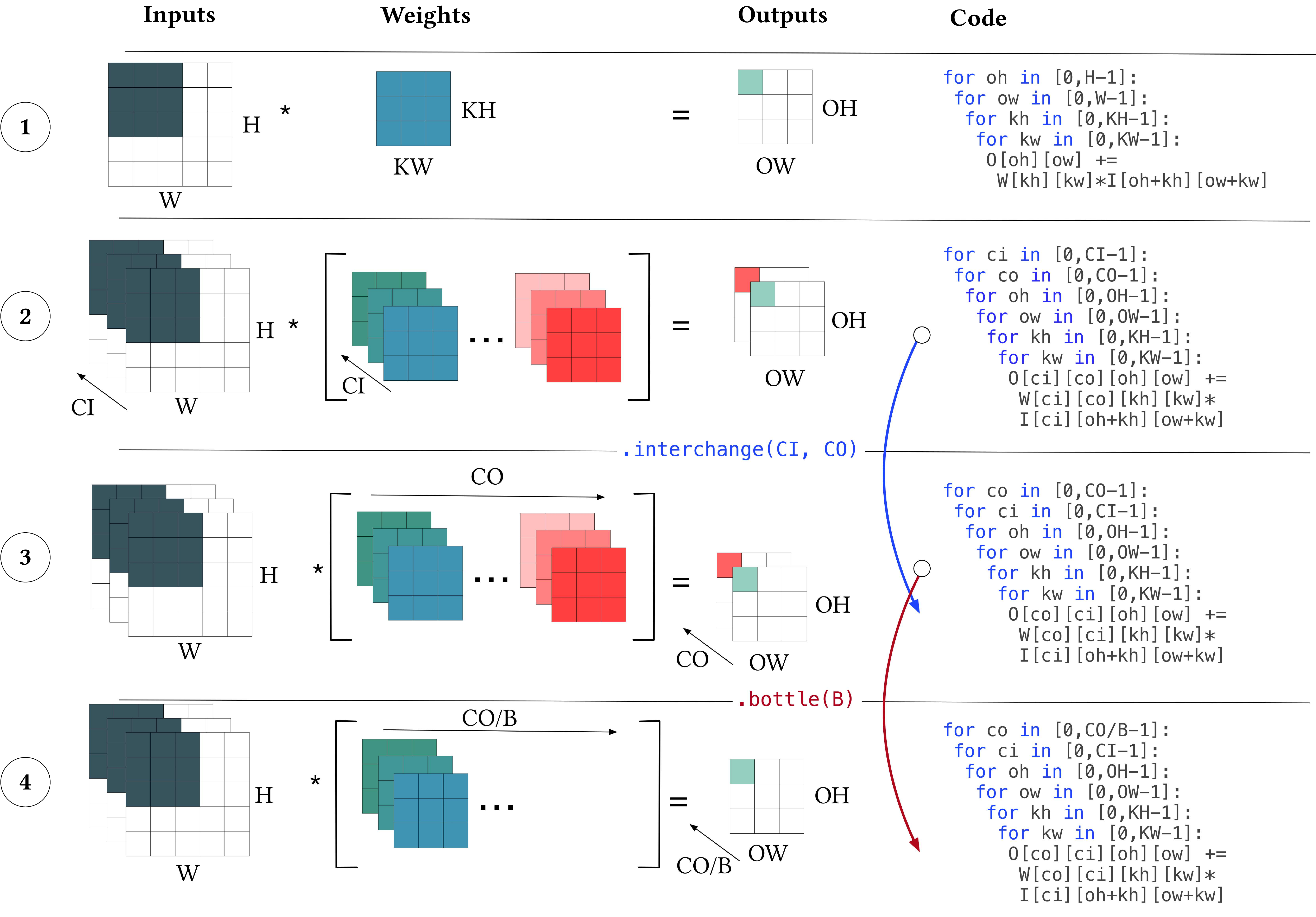}
    \caption{An illustration of the relation between neural architecture components, code and transformations. The first row shows a basic convolution. The second shows a tensor convolution. In the third row, we see a program transformation: loop interchange. The fourth row shows a neural architecture transformation: bottlenecking. Here, the number of weights is divided by $B$. The accuracy of the network after training does not change, but the number of operations being performed is drastically reduced.}
    \label{fig:conv}
\end{figure*}

\section{Overview}

\label{sec:overview}
Here we develop a simple example to illustrate the direct connection between models and code, and hence, neural architecture search (NAS) and program transformation. This is followed by an outline of our approach.

\subsection{Models and Code}
The fundamental building block of a DNN is the tensor convolution, a generalization of a basic convolution.

\paragraph{Basic Convolution}
Consider the first row in Figure~\ref{fig:conv}. This shows a basic convolution where each element in the 2D output matrix is the weighted sum of the corresponding and neighboring elements of the 2D input matrix. The weights are stored in a small 2D filter or weight matrix W whose size corresponds to the size of the neighborhood.  This is shown both diagrammatically on the left and as code on the right.

\paragraph{Tensor Convolution}
A tensor convolution is a generalization of the basic convolution as shown in  row 2  of Figure~\ref{fig:conv}. The input is now a 3D tensor with a new dimension, $C_i$, referred to as the input channels. Similarly, the output is expanded by the number of output channels ($C_o$). The weight matrix is now a 4D tensor of $C_o \times C_i$ channels with two spatial dimensions (height, width).
Two new  loops scan the channels, the outermost, $C_i$ shown by an arrow,  while each of the tensors arrays have appropriate extra dimensions.

\subsection{Models and Code Transformations}
In the compiler setting, transformations change structure while maintaining meaning.

\paragraph{Code Transformation}
Loop interchange changes the order of computation and memory access
without affecting the values  computed. In row 2 of Figure~\ref{fig:conv},
interchanging the outer loop iterators
gives the code in row 3 with the  $C_o$ and $C_i$ loops interchanged.
We can represent this using polyhedral notation $[C_i, C_o] \mapsto [C_o, C_i]$
or TVM-like syntax:
{\tt .interchange(CI,CO)}.
This is shown diagrammatically with the arrow denoting the new outermost, $C_o$  loop order. From a neural architecture perspective nothing has changed; the number of computations and memory accesses remains the same, as do the values of the output matrix.  From a code perspective, the memory access behavior is significantly altered, affecting execution time.

\paragraph{Model Transformation}
A popular choice in NAS for reducing convolutional complexity is bottlenecking~\cite{tan2019mnasnet}. For a convolution with $C_o$ filters, we  choose a bottlenecking factor $B$. This reduces the size of the weight and output tensors
 and reduces the range  of the~\textit{outermost loop iterator} by a factor $B$. Again this can be represented as   $[C_o] \mapsto [C'_o<C_o/B]$ or {\tt .bottleneck(B)}. This is shown in  row 4 of  Figure~\ref{fig:conv}. From a NAS point of view, this is a standard operation reducing complexity with a minimal effect on representational capacity.  From a program transformation point of view, this is illegal as the computed values are changed.

\subsection{Combined Space}
If we consider swapping a full convolution for a reduced convolution as a  program transformation,
we can combine them. For example, we could apply loop interchange to the code in Figure \ref{fig:conv} row 4, ($[C'_o<C_o/B, C_i] \mapsto [C_i, C'_o<C_o/B]$). With $C_i$  now the outermost iterator, we can reapply bottlenecking, giving input channel bottlenecking. Such an optimization is both semantically invalid, and  also unavailable in existing neural architecture search spaces. However, given the robustness of neural networks to noise, in some specific cases it may be just as representationally preserving as output channel bottlenecking. It is only through the lens of program transformations over loop nests that we are able to automatically access such operators.

\section{Neural Architecture Search (NAS)}
\label{sec:nas_background}
In this section we briefly describe NAS, which seeks to automate the design of neural networks for given tasks and budgets. For more details, please see \cite{wistuba2019survey}.

Starting  from an overall  network skeleton, NAS  attempts to design one or more cells that slot into different locations within the skeleton. Cells are described as a DAG with nodes as intermediate feature maps (or tensors)  and edges representing possible operations (for example, convolution or tensor product). The task is then to find the best DAG (or cell) that can be slotted into the skeleton and trained on a given dataset.  Recent work takes cells as predefined options and attempts to select where best to place each of the cells in the skeleton~\cite{wu2019fbnet, Turner2020BlockSwap:}.
to match specific constraints.

\subsection{Neural Architectures Components}

The vast majority of neural architectures consist of sequences of inter-connected components. The convolution operation dominates computational complexity and is the primary component
of interest in NAS~\cite{ying2019bench,Dong2020NAS-Bench-201:}.
We  now introduce the convolution operation, and the variants that are considered as possible substitutions in our NAS baseline.

\paragraph{Standard Convolution}
Figure~\ref{fig:conv} row 2 shows the standard convolution operation, in which an input volume  of size $H_{i} \times W_{i} \times C_{i}$ is convolved with a set of $C_{o}$ filters of size $K_h \times K_w \times C_{i}$, each producing an individual feature map in the output.
The convolution operation is
\begin{equation}
\begin{split}
    \forall c_o, w_o, h_o \hspace{2mm} O(c_o,w_o,h_o) = \hspace{2cm} \\
    \sum_{c_i}^{C_i} \sum_{k_h}^{K_h} \sum_{k_w}^{K_w} I(c_i,w_o+k_w,h_o+k_h) * K(c_o,c_i,k_w,k_h).
\end{split}
\end{equation}
\paragraph{Bottlenecked Convolution} A popular choice for reducing convolutional complexity is bottlenecking~\cite{he2016deep}
as shown in Figure \ref{fig:conv} row 4. A bottlenecking factor $B$ is selected, reducing the number of weights/filters to $C_{o}/B$. ResNets~\cite{he2016deep} frequently feature bottleneck blocks that consist of trios of convolutions, one to bring the number of channels down, another for processing, and a final one to bring the number of channels back up. A similar pattern is relied on in several NAS techniques~\cite{tan2019mnasnet, tan2019efficientnet}.
The bottlenecked convolution operation is
\begin{equation}
\begin{split}
    \forall c'_o < \frac{C_o}{B}, w_o, h_o \hspace{2mm} O(c'_o,w_o,h_o) = \hspace{2cm} \\
    \sum_{c_i}^{C_i} \sum_{k_h}^{K_h} \sum_{k_w}^{K_w} I(c_i,w_o+k_w,h_o+k_h) * K(c'_o,c_i,k_w,k_h).
\end{split}
\end{equation}

\paragraph{Grouped Convolution}
Here, the $C_{i}$ channel input is split along the channel dimension into $G$ groups, each of which has $C_{i}/G$ channels. Each group is independently convolved with its input split, producing $C_{o}/G$ channels which are concatenated along the channel dimension.
Let $O=[O'_1;\ldots, O'_G]$ i.e. each slice  $O'_g, g \in 1,\ldots G$
is concatanated to form $O$.  $I'_g$ is similarly defined.
Group convolution is then as follows:
\begin{equation}
\begin{split}
    \forall c'_o, w_o, h_o \hspace{2mm} O(c'_o,w_o,h_o) = \hspace{2cm} \\
    \sum_{c'_i}^{C'_i} \sum_{k_h}^{K_h} \sum_{k_w}^{K_w} I(c'_i,w_o+k_w,h_o+k_h) * K(c'_o,c'_i,k_w,k_h).
\end{split}
\end{equation}

This reduces the number of basic convolutions from $C_o \times C_i$ to $G \times C_o /G \times C_i /G  = (C_o \times C_i)/G$ reducing the number of operations used by a factor of $G$. Note that we can think of standard convolution as grouped convolution with $G=1$.

\begin{figure}[t]
  \includegraphics[width=\linewidth]{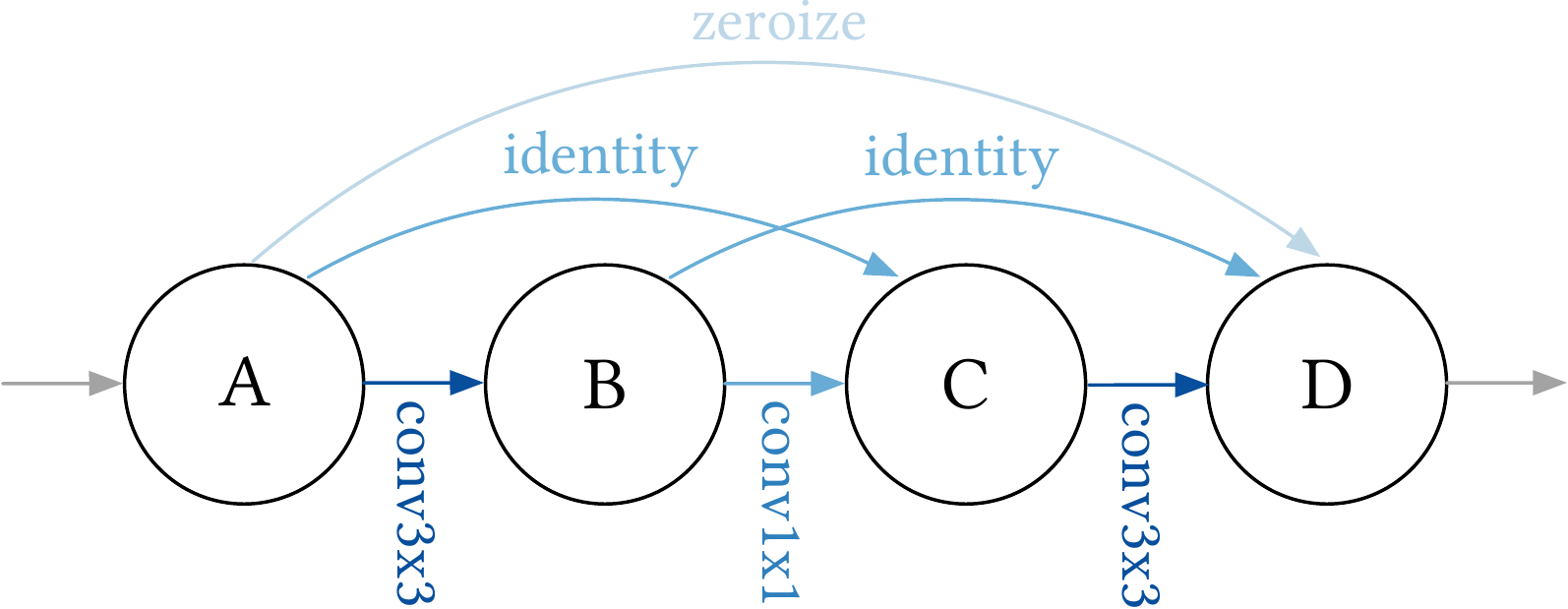}
  \caption{An illustration of a Neural Architecture Search design space (as used in~\cite{ Dong2020NAS-Bench-201:}). Every cell has exactly four nodes (A,B,C,D) representing intermediate feature map states. Edges represent operations that transform intermediate states from source to target node, taken from a list of options specified by the designer.
}
  \label{fig:nas-dag}
\end{figure}

\paragraph{Depthwise Convolution}
Notice that if the number of groups is equal to the number of input channels and the number of output channels, $G=C_{i}=C_{o}$, then there is a single, 2D convolutional filter for each input channel $C_i$.  This is known as depthwise convolution and is a special case of group convolution.

\subsection{NAS Search Space Example}
Typically, the NAS heuristic finds the best cell to fill in the skeleton by choosing appropriate operations. Figure~\ref{fig:nas-dag} gives an example cell from~\cite{Dong2020NAS-Bench-201:}. The skeleton is ResNet-like with 5 cells in series. Each cell contains 4 inter-connected nodes (A,B,C,D). Between nodes, downsampling takes place where the spatial dimensions are halved in size while  doubling the channel depth. This restricts the types of operations available on each edge.
This gives a total of 15625 possible %DAGs or
cells, which captures most of the available options within cell-based NAS techniques, each of which the authors train exhaustively on various datasets.

Rather than designing cells by choosing which of a predetermined list of options is best, we instead wish to choose a sequence of transformations from an original model which will allow us to step through the cell design space. We describe the program transformations we use in the next section.

\section{Program Transformations}
\label{sec:PT_background}
Due to the restricted, static,  convex and  affine nature of tensor convolutions,  it is natural to  describe program transformations of them in the  well-studied polyhedral model.
In this section we give a brief introduction to the model; for more a detailed description, we refer the reader to \cite{verdoolaege2014schedule,vasilache2018tensor}.
The polyhedral model of a program consists of three main components:

The~\textbf{domain} is a collection of the possible statement instances that occur within the iteration space  of a set of loop bounds.  We can represent each statement instance with a multidimensional co-ordinate corresponding to the iterator values of the loops that surround it.
As the constraints on the loops are affine, the  bounded domain forms a convex polyhedron.

A~\textbf{set of accesses} are affine mappings  of the iteration space to memory. Two statement instances have a dependence ordering between them if they have accesses to the same memory location and at least one of them is a write.

A~\textbf{schedule} assigns a timestamp to each statement instance, dictating the order in which they are executed. Different schedules for the same program represent possible program transformations. The transformed schedules are determined to be semantically preserving if dependence ordering is preserved.

\begin{algorithm}[t]
\begin{minted}[linenos,style=xcode]{C++}
for (co=0; co<Co; co++)
  for (oh=0; oh<OH; oh++)
    for (ow=0; w<OW; ow++)
S1    O[c_o][h][w] = 0.;
      for (ci=0; ci<Ci; ci++)
S2      O[co][oh][ow] +=
          W[co][1][1] *
          I[ci][oh][ow];
\end{minted}
\caption{Naive implementation of
$1 \times 1$ tensor convolution.
}
\label{alg:naive_res}
\end{algorithm}

To illustrate this, consider the implementation of the $1\times1$ convolution at the start of a residual block listed in Algorithm~\ref{alg:naive_res}.
We can describe the domain as follows:
\[
  \begin{array}{l}
 \begin{array}{l@{~}c@{~}l}
%                           \{
                           \mathtt{S1}(c_o,h,w)       & \mid & 0 \leq c_o < C_o \wedge 0 \leq h < H \wedge  \\ & & 0 \leq w < W
%                           \}
                           \\
%                           \{
                           \mathtt{S2}(c_o,h,w,c_i)   & \mid & 0 \leq c_o < C_o \wedge 0 \leq h < H \wedge  \\ & &  0 \leq w < W \wedge 0 \leq c_i < C_i
%                           \}
                           \\
 \end{array}
 \\
  \end{array}
 \]
We can also describe the schedule as follows:
\[
  \begin{array}{l}
    \quad\quad\quad
    T_{\mathtt{S1}}(c_o,h,w) = (c_o,h,w)
    \\
    \quad\quad\quad
    T_{\mathtt{S2}}(c_o,h,w,c_i) = (c_o,h,w,c_i)
    \\
  \end{array}
 \]
We now briefly discuss the transformations we consider.

\paragraph{\textbf{Loop Interchange}} Interchanging two nested loops involves applying a permutation to  the schedule of the enclosed statements. For example, in the algorithm shown in \ref{alg:naive_res} and a statement $S1$ we can express this as simply as:
\[
    T_{\mathtt{S1}}(c_o,h,w) = (c_o,w,h) %\}
\]

\paragraph{\textbf{Strip-Mining}} This  is performed by mapping an  iterator into two new iterators whose combined range is  that of the original. A constant  strip-mining factor is selected which forms the range of the new inner loop. The  outer loop range is that of the original divided by the strip-mine factor. For example, to strip mine the inner $c_i$ loop of Algorithm~\ref{alg:naive_res}  we have
\[
    T_{\mathtt{S2}}(c_o,h,w,c_i) = (c_o,w,h,c_i/32,c_i \Mod{32})
\]
\paragraph{\textbf{Tiling}} Tiling is a combined transformation
consisting of strip-mining followed by interchange.  For example, to tile  $c_i$ loop of \ref{alg:naive_res}  we have
\[
    T_{\mathtt{S2}}(c_o,h,w,c_i) = (c_i/32,c_o,w,h,c_i \Mod{32})
\]
There are many more transformations that can be expressed in the polyhedral model. For further detail, we point the reader to the excellent polyhedral literature~\cite{grosser2011polly,vasilache2006polyhedral}.
\subsection{Legality}
Legality in the polyhedral framework is determined by preservation of data dependences. If there is a data dependence between two dynamic statement instances under the original program  schedule, the relative ordering  between these statements must be preserved under the new transformed schedule.

In the case of Algorithm~\ref{alg:naive_res}, we can see that $S1$ is the definition, or source, of a dependence that must occur before its usage (in $S2$). For an instance $i$ of statement $S1$ and $j$ of statement $S2$, assuming constant data dependences, we can represent dependences in matrix form $D$. Elements of $D$ are non-negative integers such that an element $d_{S1, S2}$ indicates that $i$ must execute before $j$ (formally, $i \prec j$, where $\prec$ denotes lexicographical ordering).

Given a linear transformed schedule  of the iteration space $I$  in matrix form $T$ and a dependence matrix $D$, then a transformation is legal iff:
\[
\forall  i,j,S1,S2,D  \quad i \rightarrow j \in d_{S1,S2} \rightarrow T(i) \preceq T(j)
\]
which is to say that the lexicographical ordering is preserved.

\section{Unified Space}
\label{sec:unified}

Our approach to joining  NAS and compiler optimizations is to describe
convolutional alternatives as polyhedral program transformations. Due to the
affine nature of convolutions, this is relatively  straightforward, except for
checking transformation legality.

We adapt a newly developed legality check based on Fisher
information~\cite{Turner2020BlockSwap:}, called Fisher Potential, that allows us
to check the legality of transformations without having to retrain the network
each time. This leads to a dramatic reduction of transformation search time as
shown in Section~\ref{sec:results}.

\subsection{Extending the Polyhedral Model}
\label{section:extending}

\paragraph{\textbf{Bottlenecking}} The bottleneck transformation is a reduction
in the outer iterator in the domain node by parameterized constant factor $B$.
Let the vector $\underline{J}$ denote the iterators spanning the domain of the
tensor loop nest, $\underline{J} = [c_o,c_i,h,w,k_h,k_w]$ where $c_o$ is the
outermost iterator and $\underline{J'} = [c_i,h,w,k_h,k_w]$ be the enclosed
inner iterators, then bottlenecking can be expressed as
\[
    T_{\mathtt{S}}(c_o,J') = (c'_o,J') \mid c'_o <C_o/B
\]
The value $B$ is constrained such that $C_o \Mod{B} \equiv 0$. This gives a factor
$B$ reduction in computation. Figure 1 row 4 shows an example of bottlenecking.

\paragraph{\textbf{Grouping}}

\begin{algorithm}[t]
\begin{minted}[linenos,style=xcode]{C++}
for (g=0; g<G; g++)
  for (co=Co/G*g; co<Co/G*(g+1); co++)
    for (ci=Ci/G*g; ci<Ci/G*(g+1); ci++)
      for (oh=0; oh<OH; oh++)
        for (ow=0; w<OW; ow++)
          for (kh=0; kh<KH; kh++)
            for (kw=0; kw<KW; kw++)
              O[co][oh][ow] +=
                        W[co][ci][kh][kw] *
                        I[co][oh+kh][ow+kw];
\end{minted}
\caption{Grouping transformation of tensor convolution.}
\label{alg:group}
\end{algorithm}

Grouping can be thought of as tiling the two outer iterators by a common factor
and then discarding one of the iterators. The original iterator domains must
both be divisible by the grouping factor $G$. Let $J' = [c_i,J'']$, such that
$J= [c_o,c_i, J'']$ then we can define grouping as
\[
    T_{\mathtt{S}}(c_o,c_i,J'') = (g,c_o/G,c_i/G,J')
\]
to give the algorithm in Figure \ref{alg:group}. Note as $C_o,C_i$ and $G$ are
compile-time known constants each of the loop bounds is affine. Examining the
code we see that each slice $g$ of the output array refers only to the
corresponding slice of the weight and input array.

\paragraph{\textbf{Depthwise}}

\begin{algorithm}[t]
\begin{minted}[linenos,style=xcode]{C++}
for (g=0; g<Co; g++)
  for (oh=0; oh<OH; oh++)
    for (ow=0; w<OW; ow++)
      for (kh=0; kh<KH; kh++)
        for (kw=0; kw<KW; kw++)
          O[g][oh][ow] +=
                    W[g][kh][kw] *
                    I[g][oh+kh][ow+kw];
\end{minted}
\caption{Depthwise transformation of tensor convolution.}
\label{alg:depthwise}
\end{algorithm}
Depthwise convolutions can be considered as a special case of group convolutions
where the group size equals the number of output channels, $C_o, G=C_o$. For
this transformation to be possible, the number of input and output channels must
be equal, $C_o=C_i$ . This means the two inner loops will have strip counts of 1
as $C_o/G= C_i/G = 1$ and
\[
    T_{\mathtt{S}}(c_o,c_i,J'') = (g,1,1,J')
\]
which can be trivially simplified to
\[
    T_{\mathtt{S}}(c_o,c_i,J'') = (g,J')
\]
\subsection{Fisher Potential as a Legality Check}
\label{sec:fish}

The transformations described above fail to preserve traditional program semantics.
We instead state that a schedule transformation for a neural network with
respect to a task is legal if the final classification accuracy on the held out
data is the same, or similar to within a small~$\delta$. Training all possible
transformed networks to determine accuracy, however, would be prohibitively
expensive.

To address this, we employ~\textit{Fisher Potential}~\cite{Turner2020BlockSwap:}
as a pre-training legality check. It is a cheap-to-compute measure that is
effective at rejecting any architectures whose final test accuracy is
significantly below the original starting point, {\em without needing to perform
training}. Informally, Fisher Potential is, locally, the total information that
each loop nest (layer) contains about class labels under a simplifying
assumption of conditional independence. We note that the specific measure
could easily be swapped out for another, such as~\cite{mellor2020neural},
or any of the measures developed in~\cite{abdelfattah2021zero}, and
that this is an area of active development.

Fisher Potential refers to the use of aggregated Fisher Information at
initialization to estimate the effectiveness of network architectures before
training. Where~\cite{lecun1989optimal} used the diagonal of the Hessian in
order to compress networks,~\cite{molchanov2017pruning,theis2018faster} showed
that this computation could be approximated via the Fisher Information Matrix.
This approximation was empirically shown to be effective at estimating the
importance of individual neurons in a neural
network~\cite{molchanov2019importance}, and to correlate to final
accuracy~\cite{golatkar2019time}. In SNIP~\cite{lee2019snip}, the authors used
a measure heavily related to Fisher Information~\textit{at initialization} in order to prune connections
in the network prior to the training process. This was adapted
in~\cite{Turner2020BlockSwap:} to perform architecture search, by summing the
Fisher Information over neurons to get layer-wise importance estimates. It is
this form of the measure that we use in this paper.

\paragraph{Formal Definition}
For a single channel, $c$, of a convolutional filter in a network, consider that
for some input minibatch of $N$ examples, its outputs are an $N \times W \times
H$ tensor where $W$ and $H$ are the channel's spatial width and height. We refer
to this tensor as the activation $A$ of  this channel and denote the  entry
corresponding to example $n$ in the mini-batch at location $(i,j)$ as $A_{nij}$.
As the network has a loss function $\mathcal{L}$, then we can get the gradient
of the loss with respect to this activation channel $\frac{\partial
\mathcal{L}}{\partial {A}}$. Let us denote this gradient as $g$ and index it as
$g_{nij}$. The channel $c$ error,  $\Delta_{c}$ can then be computed by
\begin{equation}
 \Delta_{c} =\frac{1}{2N} \sum_{n}^{N}\left(- \sum_{i}^{W}\sum_{j}^{H}A_{nij} g_{nij}\right)^2. \label{eqn:fisher}
\end{equation}
This gives us a filter-wise score for a particular channel. In order to gauge
the sensitivity of the full convolution, we sum $\Delta_{c}$ over each channel
(as in~\cite{Turner2020BlockSwap:}):
\begin{equation}
    \Delta{l} = \sum_{c_o}^{C_o} \Delta_{c_o}
\end{equation}
This score is summed for each of the convolutional blocks in the network. For an
original network and a proposed alternative architecture, we reject the proposal
if its score is below that of the original.

To summarise, the Fisher Potential of a proposed network architecture is the sum
of the Fisher Information for each layer when given a single random minibatch of
training data, as performed in~\cite{Turner2020BlockSwap:}.

\begin{figure}[t]
\includegraphics[width=\linewidth]{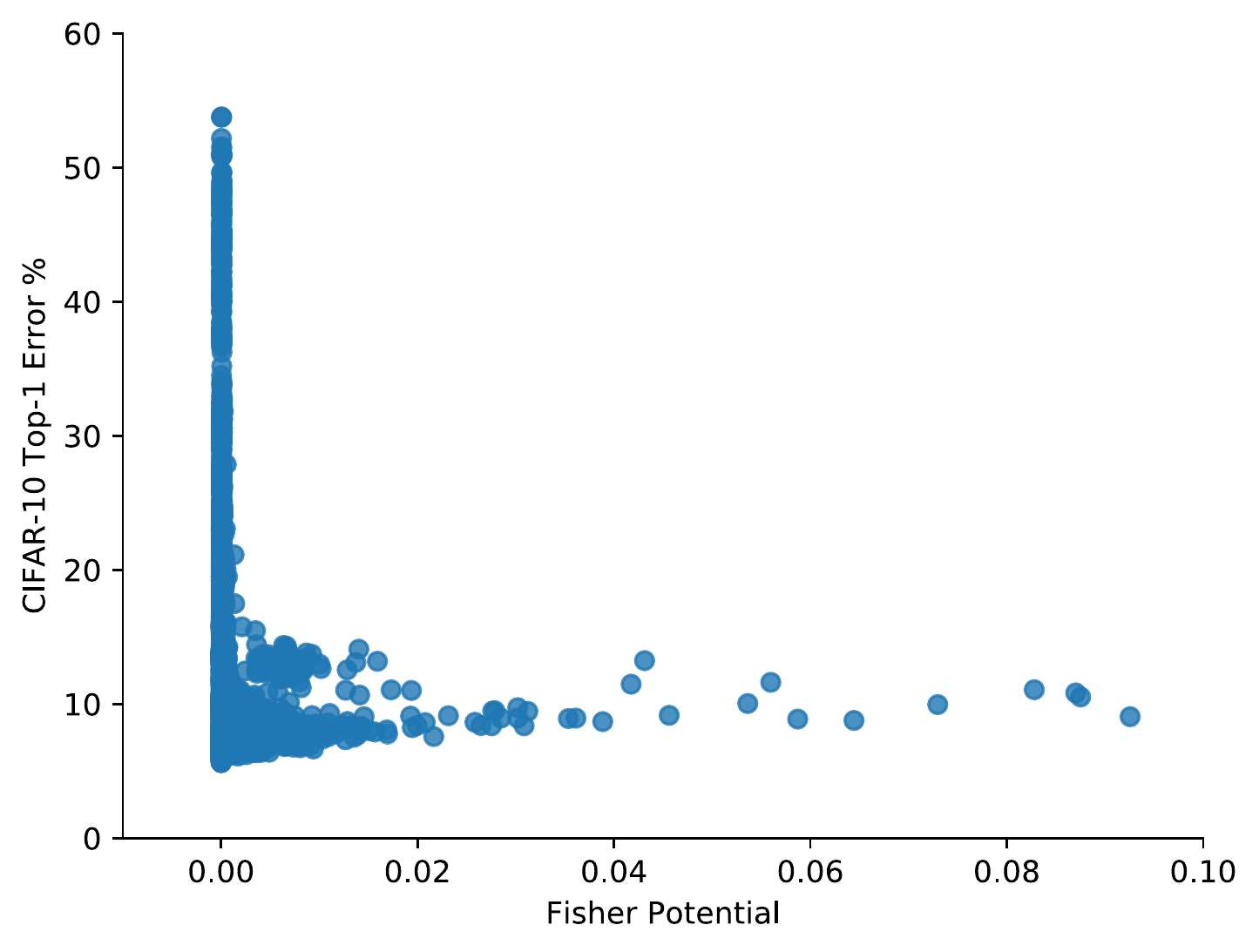}
\caption{Fisher Potential as a rejection filter for invalid architectures. Each
point is a different neural architecture from
NAS-Bench-201~\cite{Dong2020NAS-Bench-201:}. The $y$-axis shows final CIFAR-10
top-1 error (lower is better), and the $x$-axis shows the Fisher Potential
assigned to the model~\textit{at initialization} (higher is better).}
\label{fig:fishpo}
\end{figure}

\paragraph{Example}
Let us consider, candidate networks designed from
NAS-Bench-201~\cite{Dong2020NAS-Bench-201:} as shown in Figure \ref{fig:fishpo}.
Each point represents a different neural architecture of which there are 15625.
The $y$-axis shows final CIFAR-10 top-1 error (lower is better), and the
$x$-axis shows the Fisher Potential assigned to the model~\textit{at
initialization} (higher is better). We can see that without requiring any
training, Fisher Potential is able to filter out poorly-performing
architectures, visible in the cluster of low scoring networks on the left with
poor final errors. Many good networks are also discarded --- this is unfortunate but
acceptable for our scenario, since the space is large and densely populated with
networks. We note again that this measure in particular could be swapped out for an
improved one in future.

\subsection{Expressive Power}
\label{sec:spatial}
Neural Architecture Search techniques rely on hand-designed exploration spaces.
themselves are engineered by hand.
For example, a recent paper found that bottlenecking could be applied in the
spatial domain~\cite{peng2018accelerating}. This can now be added to a list of
candidate operations.
However, spatial bottlenecking  is automatically captured within our framework
as the combination of existing transformations. As an example, consider the
convolution in row 2 of Figure 1. We use the shorthand notation $[i]
\xrightarrow[]{B(b)} [i(b)]$ to denote bottlenecking domain $i$ by factor $b$
and $[i,j] \xrightarrow[]{\text{int}} [j,i]$ to denote interchange. Then the
spatial bottleneck operation can be constructed as the following sequence of
transformations:

$\displaystyle
  \begin{array}{l}
   T_S: [C_o, C_i, H, W, K_h, K_w] \xrightarrow[]{\text{int.}} \\
  \quad\quad [H, W,C_o, C_i, K_h, K_w] \xrightarrow[]{\text{B(b).}} \\
  \quad\quad [H(b), W,C_o, C_i, K_h, K_w] \xrightarrow[]{\text{int.}} \\
  \quad\quad [W,H(b),C_o, C_i, K_h, K_w] \xrightarrow[]{\text{B(b).}} \\
  \quad\quad [W(b),H(b),C_o, C_i, K_h, K_w] \xrightarrow[]{\text{int.}} \\
  \quad\quad [C_o, C_i, H(b), W(b), K_h, K_w]
  \end{array}
$
\vspace{.3cm}

The expressiveness of the polyhedral model, equipped with these new transformations, means that there are novel transformations that can be derived from this unified framework.

\section{Implementation and Setup}
\label{sec:transf_impl}
We  expressed our new transformations as polyhedral transformations and  implemented the resulting operators in TVM~\cite{chen2018tvm}. We selected TVM as its API allows composition of transformation sequences, and is a well-accepted state-of-the-art optimizing compiler for convolutional neural networks.

\paragraph{Transformation Space}

An overview of the transformations used in this paper is given in Table~\ref{tab:opt-space}.
The first four are standard program transformations available within TVM.
To these, we add our two neural architecture transformations. Finally, for GPU platforms, we enable GPU mapping transformations.

\paragraph{Baseline TVM}
TVM allows  users  to implement schedules for each new operator by hand. Due to the large
number  of operators involved,  for each device we use TVM's default schedules;
automatic schedule design~\cite{zheng2020ansor} has yet to be incorporated into TVM.
We then enable  auto-tuning of parameter values within the  schedule to find best performance.

\paragraph{Search}
Our current search process is relatively naive. We enumerate random sequences of transformations through TVM, and generate 1000 configurations of the resulting operations for each network.
We then check which candidates pass our implementation of the Fisher Potential legality test and select the best performing one.

\begin{table}[t]
\centering
\caption{A description of the autotuning primitives available to TVM for optimizing operations, including GPU mapping, standard primitives, and our two new optimizations.}
\begin{tabular}{|p{2.3cm}|p{5.4cm}|}
\hline
    \textbf{Optimization} & \textbf{Description} \\
\hline
 \multicolumn{2}{|c|}{Program Transformations}\\
\hline

\texttt{reorder}      &  Interchange nested loops \\
\texttt{tile}         &  Cache and register blocking     \\
\texttt{unroll}       &  Loop unrolling  \\
\texttt{prefetch}     &  Memory coalescing between threads \\
\texttt{split}        &  Divide iteration into multiple axes \\
\texttt{fuse}         &  Combine two axes into one \\
\hline
  \multicolumn{2}{|c|}{Neural Architecture Transformations} \\
\hline

\texttt{bottleneck} &  Reduce domain by factor $B$ \\
\texttt{group}      &  Slice and offset two loops by factor $G$ \\

\hline
 \multicolumn{2}{|c|}{Mapping to GPU} \\
\hline
\texttt{blockIdx}     & Block-wise parallelism \\
\texttt{threadIdx}    & Threads within blocks \\
\texttt{vthread}      & Striding thread access \\

\hline

\end{tabular}

\label{tab:opt-space}
\end{table}

\subsection{Experimental Setup}
\label{sec:setup}

\paragraph{Platforms}
We evaluate on  an ARM A57 mobile CPU (mCPU), and an Nvidia 128-Core Maxwell mobile GPU (mGPU) available on the Jetson Nano board. We also evaluate  an Intel Core i7 (CPU) and an Nvidia 1080Ti (GPU). This is representative of a wide range of deployment targets, from mobile to server class. We use TVM~\texttt{v1.7} compiled with CUDA 10.1 and LLVM 8.0. For training and accuracy evaluation we use PyTorch~\texttt{v1.4}.

\paragraph{Neural Models}
We evaluate our methodology on three popular networks: ResNet-34~\cite{he2016deep}, ResNeXt-29~\cite{xie2017aggregated}, and DenseNet-161~\cite{huang2017densely}. These networks were chosen to represent a range of convolutional architectures, from standard $3\times3$ convolutions in ResNet-34 to grouped convolutions in ResNeXt and a heavy reliance on $1\times1$ convolutions in DenseNet.
CIFAR-10 models are trained for 200 epochs with Stochastic Gradient Descent (SGD)~\cite{bottou2010large} and a learning rate of 0.1, decayed by a factor of 10 at epochs 60, 120, and 160. ImageNet models were trained using the default PyTorch script\footnote{https://github.com/pytorch/examples/tree/master/imagenet}, which trains for 90 epochs with SGD, starting from a learning rate of 0.1 and decaying by a factor of 10 every 30 epochs.

\paragraph{Comparison}
The models are implemented with each operation written as a TVM Tensor Expression~\cite{chen2018tvm}, which is an einsum-style syntax for expressing tensor computations.
This is lowered to TVM IR, where our transformations can be employed. This allows for a fair comparison of each approach.

For each network, we consider three approaches. First we compile the model using
TVM using its default schedule (labeled TVM). We report the best performance
achieved after auto-tuning. Next, we use BlockSwap~\cite{Turner2020BlockSwap:}
as NAS to compress the modifiable convolutions in the network, followed by compilation with
TVM (labeled NAS). Finally, we apply our unified approach as described in the
previous section (labeled Ours).

\begin{figure*}[t]
\begin{tabular}{c}
  \includegraphics[width=.9\linewidth]{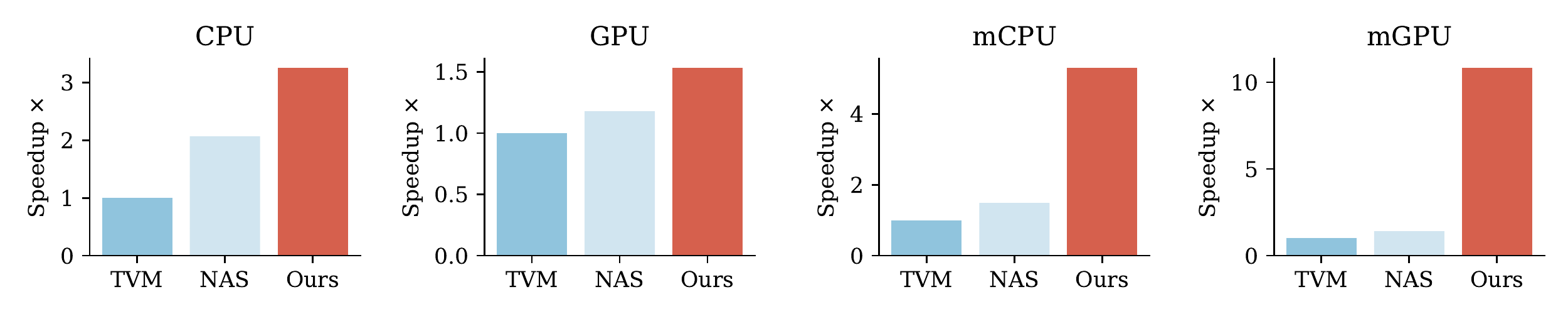} \\
  (a) ResNet-34 \\
  \includegraphics[width=.9\linewidth]{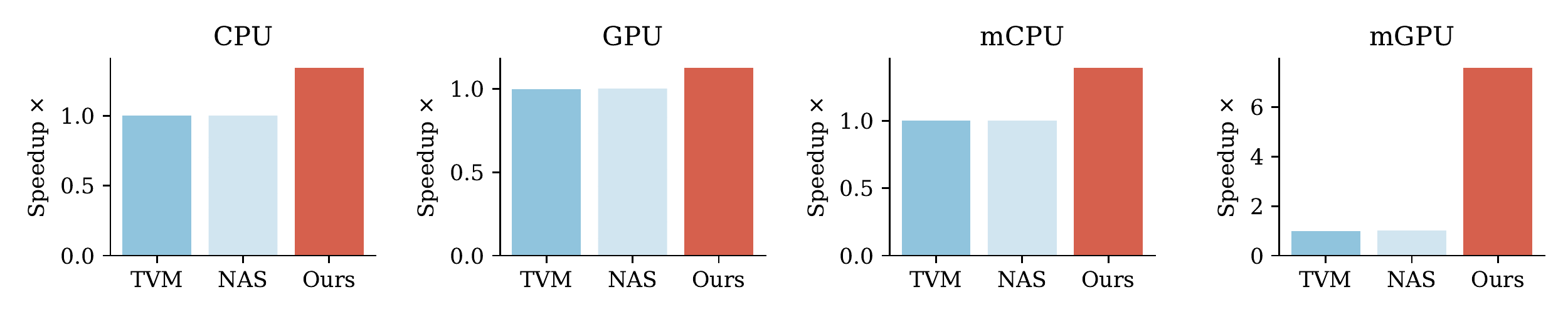} \\
  (b) ResNext-29-2x64d \\
  \includegraphics[width=.9\linewidth]{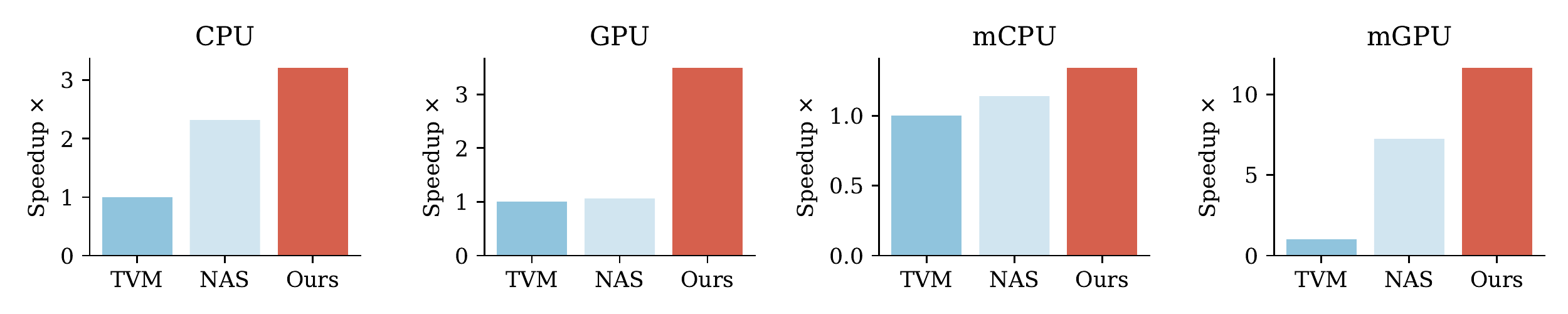}\\
  (c) DenseNet-161\\
\end{tabular}
\caption{End-to-end performance for several networks on different hardware devices on CIFAR-10. TVM represents the original model compiled with TVM default schedules. The NAS columns represent the BlockSwap-compressed copies of the models
which are then compiled with TVM default schedules. Ours represents our unified NAS-compilation strategy, with new operators stacked into the network blocks.}
\label{fig:results}
\end{figure*}

\begin{figure}
\includegraphics[width=.8\linewidth]{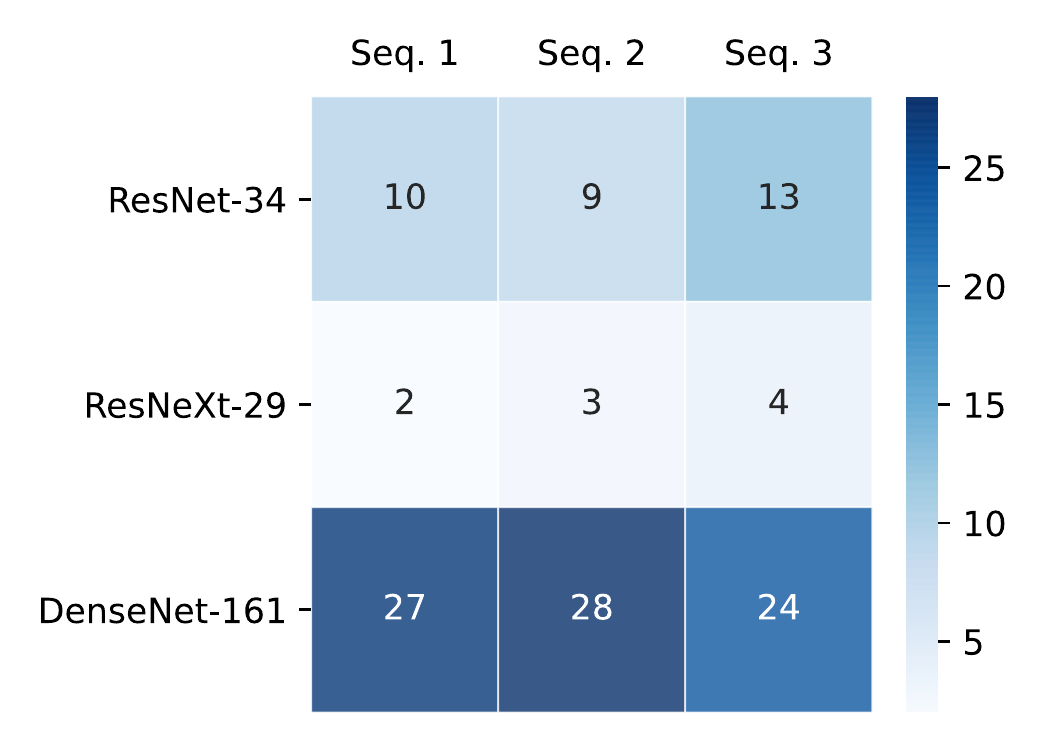}
\caption{Frequency of operation application.}
\label{fig:opfreq}
\end{figure}

\begin{figure*}
 \includegraphics[width=0.9\linewidth]{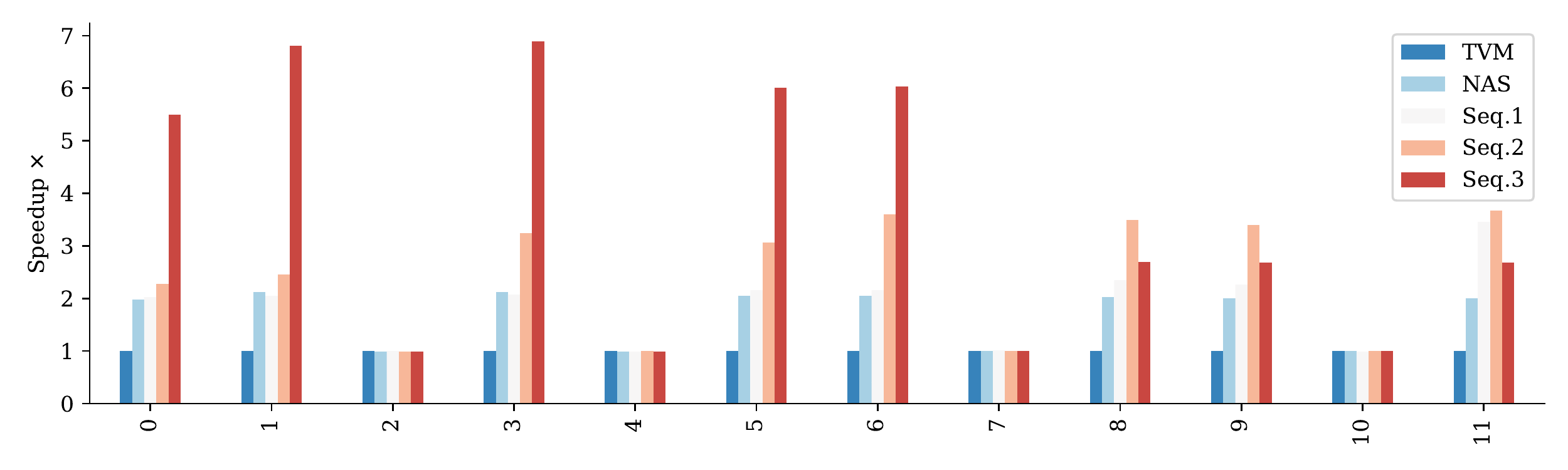}
\caption{Exploring different sequences of transformations for an individual layer of ResNet-34 on the Intel Core i7 CPU.
~\texttt{NAS} is the result of applying grouping with factor 2 first, then compiling with TVM. The other three sequences are interleaved transformations produced by our method.}
\label{fig:resbench}
\end{figure*}

\begin{figure}
  \includegraphics[width=\linewidth]{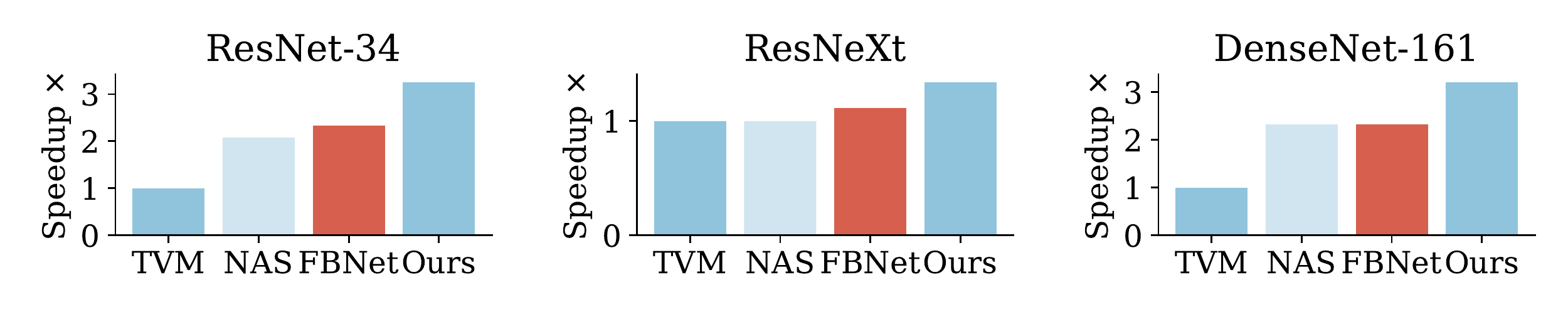}
  \caption{Intel Core i7 performance of FBNet on the three networks. FBNet is able to improve over NAS with large training cost. Our approach outperforms FBNet with no training required }
\label{fig:more_nas}
\end{figure}

\begin{figure*}
  \includegraphics[width=\linewidth]{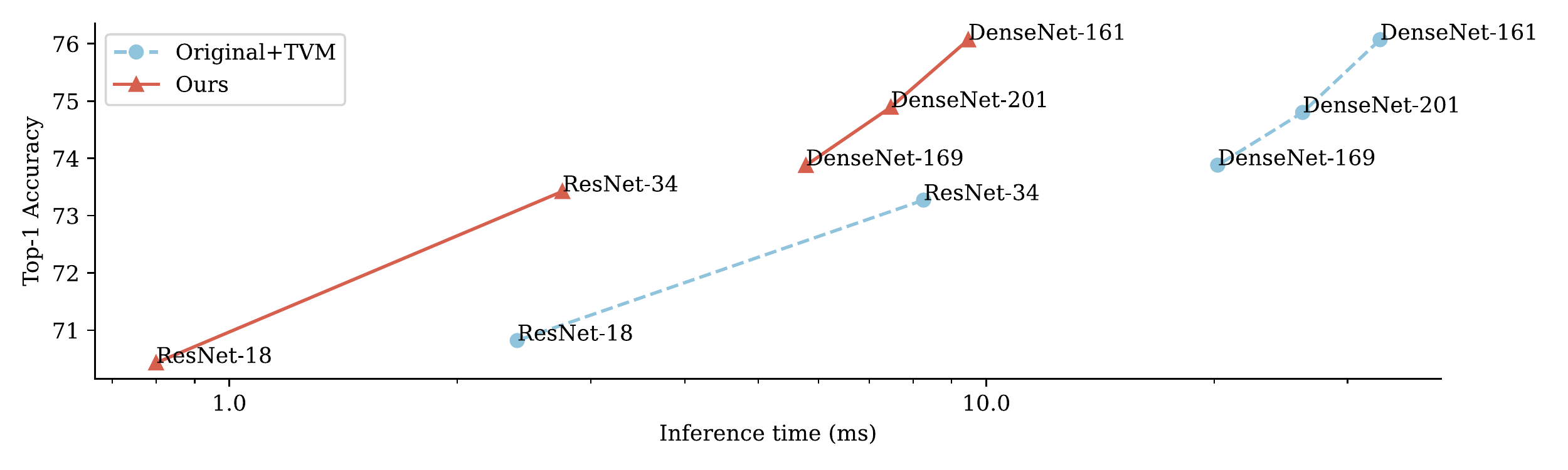}
  \caption{Accuracy vs inference time (log scale)  of our approach (Ours) compared to TVM (Original + TVM) when applied to  different variants of ResNet and DenseNet on the ImageNet dataset. Our approach gives a significant reduction in inference time.}
  \label{fig:imagenet}
\end{figure*}

\begin{figure}
  \includegraphics[width=\linewidth]{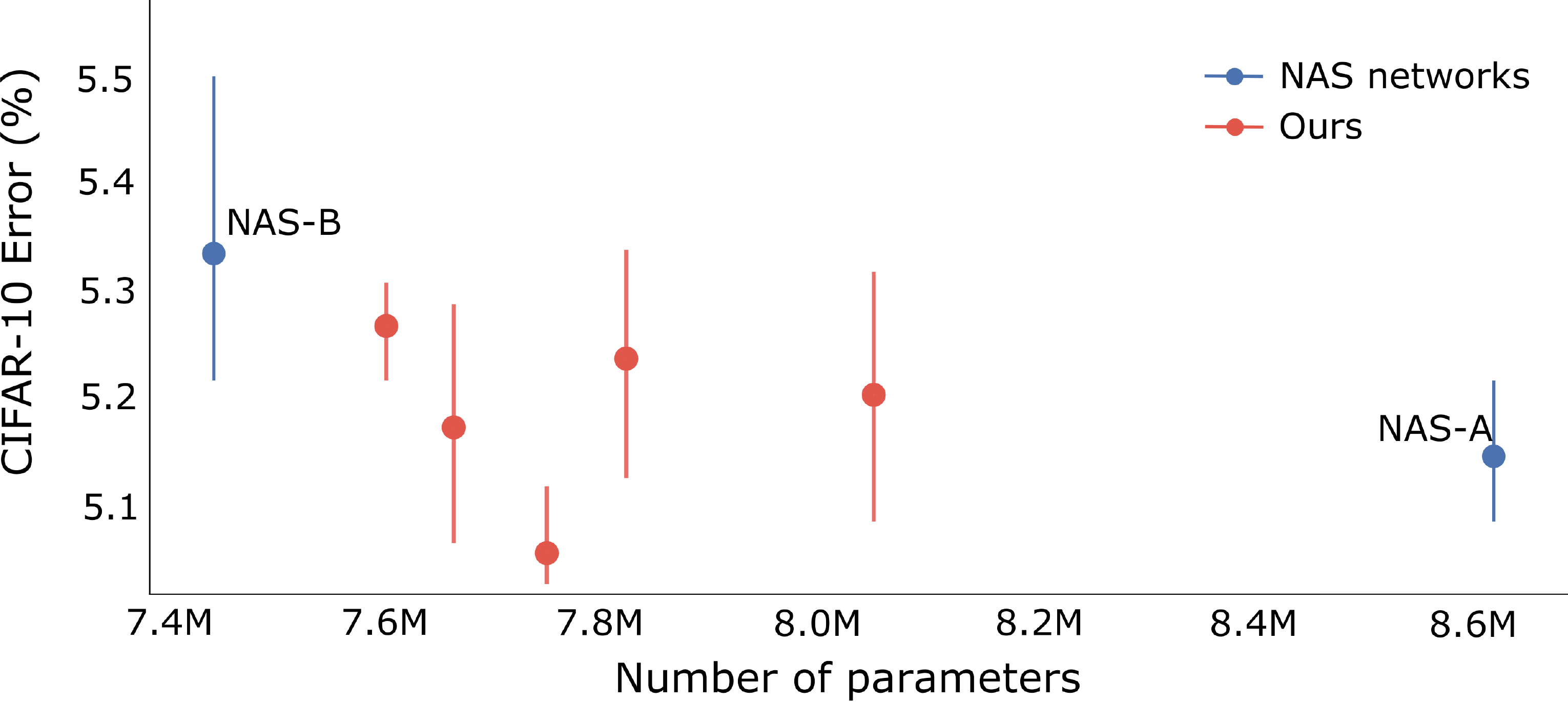}
  \caption{Two NAS models (the blue points labeled NAS-A and NAS-B) composed of grouped blocks (with g=2 and g=4 respectively) can be chained together by a series of parametrized transformations in our framework, yielding the points in red. Each point is the mean of three training runs, with error bars. }
  \label{fig:g2_g4}
\end{figure}

\section{Results}
\label{sec:results}

In this section we first present the performance of the three approaches on different networks and  platforms for  the CIFAR-10 dataset. This is followed by an analysis of accuracy, size, search time and the highest performing new convolutions found.   We then drill down into one network and
examine the impact of transformations layer-by-layer.
In order to compare  against alternative NAS techniques, we evaluate the performance of FBnet across the  three networks.
Next, we  evaluate the performance of our approach on the ImageNet dataset additionally examining the tradeoff in accuracy and performance across variants of the ResNet and DenseNet family of networks. Finally we show that our approach allows for fine-grained exploration of an accuracy/size tradeoff, producing a new Pareto optimal network design.

\subsection{CIFAR-10 Network Performance}

The results for each network are shown in Figure~\ref{fig:results}. The combination of TVM and NAS is a strong baseline, however, for each of the models our unified method is able to generate models with improved hardware performance. In general there is more performance to gain on the mGPU than other platforms, as relaxed memory pressure from smaller designs is of increasing importance.

\paragraph{ResNet-34}
NAS is able to find a modest improvement of 1.12$\times$ speedup over TVM on the GPU platform, increasing to 2$\times$ on i7 CPU, showing the power of compressed convolutions.
Our unified approach  is able to find further improvement giving 2$\times$ and 3$\times$ speedup respectively. The improvement is more dramatic on the smaller platforms with 5$\times$ and 10$\times$ speedups on the mCPU and mGPU.

\paragraph{ResNext-29-2x64d}
NAS is unable to find any improvement here due to the already highly compact structure of the network. There are simply no NAS options that improve performance over the TVM
baseline across all platforms.  Despite this our approach is able to find  small improvements of 1.3$\times$ and
1.1$\times$ on the CPU and GPU platforms by combining NAS and program transformations. This increases to 1.4$\times$ on the embedded CPU and 7$\times$ on the mGPU.

\paragraph{DenseNet-161}
The impact of NAS is much more varied here. While it can find 2.2$\times$ improvement on the CPU,
it has a negligible improvement over TVM on the GPU. It also struggles on the mCPU but finds 6$\times$ on the mGPU. Our approach is able to improve by over 3$\times$ for both server platforms, but only finds 1.2$\times$ on the mCPU while acheiving 10$\times$ on the mGPU.

\subsection{Analysis}

{\bf Accuracy}
For all CIFAR-10 networks pictured in Figure~\ref{fig:results},
changes in accuracy were less than  1\% in absolute difference. The ResNet-34 in
Figure~\ref{fig:resbench} had an original ImageNet Top-1 and Top-5 accuracy of
73.2\% and 91.4\% respectively. The final network compiled with our
transformations had a Top-1 and Top-5 accuracy of 73.4\% and 91.4\%; it was
slightly more accurate than the original but considerably faster.
To give a broader overview, we present the accuracy of several ImageNet models in
Figure~\ref{fig:imagenet}, which shows that for each model accuracy
degradation is small or non-existent.

{\bf Size:} One benefit of these neural transformations is their effect on model
size, both in terms of weights and runtime memory usage. We found that CIFAR-10
networks could be compressed in size by $2-3\times$. Likewise, the ImageNet
ResNet-34 in Figure~\ref{fig:resbench} was compressed from 22M parameters to 9M
without a loss in accuracy.

{\bf Search time: } Our search process involves suggesting configurations and
rejecting or accepting them based on Fisher Potential. Since Fisher Potential is
extremely cheap to compute, the search time overhead introduced by our method is
small, less than 5 minutes on a CPU. During this time we were able to explore
1000 different configurations, discarding approximately 90\% of the candidate
transformation sequences through the Fisher Potential legality check.
\subsection{Transformation Sequence Case Studies}

There were 3 particular transformation sequences that dominated the list of best performing transformations. In neural architecture terms, the resulting operations of each of these sequences of transformations is a new convolution-like operator, previously unavailable to NAS.

~\texttt{Sequence 1} applies grouping to the kernels over the spatial domain of the input.
These are then concatenated to form one output. The exact sequence is: \texttt{[split $\rightarrow$ interchange $\rightarrow$ group $\rightarrow$ interchange $\rightarrow$ fuse]}.

~\texttt{Sequence 2} is an operator in which the output channels have been unrolled by factor $16$ and then the remaining iteration domain has been grouped by factor $G=2$.
The exact sequence is: \texttt{[unroll $\rightarrow$ group $\rightarrow$ interchange]}
~\texttt{Sequence 3} is an operator that was devised by splitting up the iteration domain of the output channels, and applying different levels of channel grouping to each new domain (e.g. $G=2$ on first half, $G=4$ on the second half).
The exact sequence is:~\texttt{[split $\rightarrow$ group $\rightarrow$ interchange $\rightarrow$ group]}.

Figure \ref{fig:opfreq} shows how often each of these sequences  appear in our best performing
networks. ResNext-29 has the fewest  instances as it contains the fewest layers, while
DenseNet-161 has the most layers and hence most instances.  Each of them is
applicable across all networks and may be more widely applicable  as standard tensor convolutions for other networks.

\subsection{Exploring Layer-Wise Optimizations}

In Figure \ref{fig:resbench}, we examine the impact of our sequences layer-by-layer on one network on one platform: ResNet-34 on the i7. The exact configurations mirror the experiment in the original TVM paper~\cite{chen2018tvm}.
While we achieve a 3x improvement overall as shown in Figure \ref{fig:results}, the speedup achieved varies considerably  across layers.
In 4 of the 11 layers no performance improvement is found, as Fisher Potential marks these individual layers to be extremely sensitive to compression.
Simple grouping with a factor $G=2$ is able to give around 2$\times$ speedup across 7 of the 11 layers.~\texttt{Sequence 1} is able to give a small improvement in  most cases --- particularly layer 11 --- where spatial reduction yields new parallelization opportunities.~\texttt{Sequence 2} provides a slight improvement in many cases, and through the later layers it is able to offer larger speedups through improved data reuse. ~\texttt{Sequence 3} is the best option for most early layers but suffers compared to other sequences in the later layers.

\subsection{Alternative NAS comparison}

To provide an alternative evaluation against an existing NAS technique, we re-implement FBNet~\cite{wu2019fbnet} using the convolutional blocks available in our NAS space, and our three baseline networks as the skeletons into which the selected cells are inserted. Figure \ref{fig:more_nas}, shows the performance of resulting networks on CIFAR-10 relative to TVM, NAS and our approach on the Intel Core i7 for each network.
In each case, FBNet does provide a modest improvement over NAS.
However, it is worth noting that this involves an expensive training step at each stage of evaluation (requiring $\sim$3 GPU days per network). Our approach is able to consistently improve over FBNet, with no training required.

\subsection{ImageNet Network Performance}

We also applied our technique to the ImageNet classification dataset, to show that the method transfers beyond CIFAR-10. In Figure~\ref{fig:imagenet}, we first show the resultant networks for running our method on ResNet-18, ResNet-34, DenseNet-161, DenseNet-201, and DenseNet-169, with their inference times recorded on the Intel i7 CPU. For each network, accuracy is within 2\%, and there are significant gains in inference time.

\subsection{Interpolating Between Models}

The additional expressive power of our method allows us to explore the search space of networks in a more fine-grained manner than traditional NAS techniques. This is because NAS techniques simply choose operations from a list, whereas we generate new ones. To illustrate this consider Figure~\ref{fig:g2_g4} where we plot the accuracy and inference time of a ResNet-34
with two BlockSwap-based models (the blue points labeled NAS-A and NAS-B). We can explore alternatives by a series of parametrized transformations in our framework, yielding the points in red. Each of these points is a new possible blocktype that would not be accessible to a traditional NAS technique unless explicitly written by the human designer.
During this interpolation we are also able to find a Pareto optimal point.

\section{Related Work}
\label{sec:related_work}
\paragraph{Compiler optimization}
There has been much interest in  autotuning DNN code
generators~\cite{jia2019taso,chen2018tvm,vasilache2018tensor,steuwer2017lift, gibson2020optimizing, mogers2020automatic}.
Polyhedral compilers are particularly well-suited
~\cite{vasilache2018tensor,zerrell2019stripe} as they have in-built abstractions
for exploiting parallelism and memory layout in a principled form. Polyhedral
compilation itself benefits from a wealth of
literature~\cite{kelly1998framework,cohen1999analyse,verdoolaege2014schedule,vasilache2006polyhedral,Girbal2006,uday08}
that describes a unified compilation scheme and  associated set of legality
checks delivering state-of-the-art performance across a wide range of standard
compiler
benchmarks~\cite{verdoolaege2010isl,grosser2011polly,verdoolaege2013polyhedral},
as well as image and neural network specific
ones~\cite{mullapudi2015polymage,vasilache2018tensor,zerrell2019stripe}.

Other popular approaches include loop synthesis~\cite{chen2018tvm} and rewrite
rules~\cite{steuwer2017lift} which optimize through parametrized schedules. In
particular, TVM~\cite{chen2018tvm} is an adaptation of
Halide~\cite{ragan2013halide} with specific extensions for deep neural network
abstractions. Triton~\cite{tillet2019triton} is a tile-based IR for
parallelizing tensor-based computation. Most of these frameworks focus on
operator-level optimization, though higher level graph transformation has also
shown promising results ~\cite{jia2019taso}.

Though the toolchains are exceptionally feature-rich, there is evidence that
some implementations have been highly-engineered for specific workloads, at the cost of general
support for newer optimizations~\cite{barham2019machine}. All current
approaches, however, are limited by their inability to exploit NAS
transformations.

Reducing the amount of computation at the cost of accuracy has been examined in
a compiler context using loop
perforation~\cite{sidiroglou2011managing,sampson2015accept,figurnov2015perforatedcnns} in approximate
computing. Within approximate computing, there is extensive work on probabilistic
program transformations~\cite{rinard2006probabilistic,misailovic2011probabilistically,zhu2012randomized}
and language support for approximation~\cite{goiri2015approxhadoop,park2015flexjava}.
Where some approximate computing methods accept small absolute differences
in the outputs of programs, our transformations may render the numerical outputs of
our programs completely different while maintaining the legality of the program.
The benefit of Fisher Potential, while it does not give any guarantees on behavior,
is that its domain specificity allows us to capture these transformations.

\paragraph{Model Optimization}
From a machine-learning perspective, deep neural networks should be highly
compressible as there is significant evidence that they are vastly
overparametrized~\cite{denil2013predicting,frankle2019lottery}. There are
several popular strategies that can be employed. One can prune a network by
removing unimportant connections between
weights~\cite{han2016deep,gale2019state,molchanov2017pruning}. However, the
sparsity introduced can lead to poor memory behavior \cite{crowley2018pruning}. An alternative is to
retain dense weight tensors by pruning channels~\cite{li2017pruning} although
the resulting irregular structures cause a significant slowdown \cite{radu2019performance}.

Network distillation~\cite{ba2014do,hinton2015distilling, turner2018distilling},  takes a trained
large network and uses its outputs to assist in the training of a smaller
network. This, however, leads to a large design space as there are many ways to
make a network smaller
~\cite{chollet2017xception,xie2017aggregated,ioannou2017deep,gao2018condensenet,howard2017mobilenets}.

~\textit{Neural architecture search} instead automates the process of finding
efficient networks. In~\cite{zoph2017neural} the authors use an RNN to generate
network descriptions and filter the options using reinforcement learning.
Several networks are generated and trained in parallel, which requires a large
quantity of possible networks to be stored at once. To reduce this overhead, an
alternative method is to design one~\textit{supernet} that contains all of the
possible subnetworks~\cite{pham2018efficient}. This allows all models to have
access to a shared set of weights, reducing the computational requirement.
Subsequent works have made extensive use of this
technique~\cite{liu2019darts,luo2018neural,chen2018searching,wu2019fbnet}.

However, there is some evidence to suggest that weight sharing schemes
aggressively hamper the ability of the NAS agent to identify good architectures,
and under such constrained spaces, random architecture search provides a
competitive baseline~\cite{li2019random,sciuto2019evaluating}. Although NAS is a
rapidly developing area, no-one to the best of our knowledge has considered
formalizing NAS as program transformation.

\section{Conclusions and Future Work}
\label{sec:conclusion}

This paper presents a new unified program transformation approach to optimizing convolution neural networks by expressing neural model operations as formal
program transformations.
We  develop a new legality check based on Fisher potential,
and  show that different types of convolution can be described as
transformations of more fundamental building blocks.
We implemented this approach in TVM and show that our combined approach significantly
outperforms both TVM   and state-of-the-art NAS.
We eliminate the need to train while searching, dramatically reducing search time.
Future work could consider further transformations in the neural architecture space in order for a more extensive analysis, and to expand the possibilities beyond that of just convolutional networks.

\begin{acks}
The material is partially based on research by Michael O'Boyle which was sponsored
by  Defense Advanced Research Projects Agency (DARPA) under agreement number FA8650-18-2-7864. The views and conclusions contained herein are those of the authors and do
not represent the official policies or endorsements, either expressed
or implied, of DARPA or the U.S. Government.
This work was partially supported by the Engineering and Physical Sciences Research Council (grant EP/L0150), EPSRC Centre for Doctoral Training in Pervasive Parallelism at the University of Edinburgh, School of Informatics.
\end{acks}

\bibliographystyle{ACM-Reference-Format}
\bibliography{acmart}

\end{document}